# A novel graph structure for salient object detection based on divergence background and compact foreground

*Chenxing Xia; Hanling Zhang; Keqin Li*

**Abstract** In this paper, we propose an efficient and discriminative model for salient object detection. Our method is carried out in a stepwise mechanism based on both divergence background and compact foreground cues. In order to effectively enhance the distinction between nodes along object boundaries and the similarity among object regions, a graph is constructed by introducing the concept of virtual node. To remove incorrect outputs, a scheme for selecting background seeds and a method for generating compactness foreground regions are introduced, respectively. Different from prior methods, we calculate the saliency value of each node based on the relationship between the corresponding node and the virtual node. In order to achieve significant performance improvement consistently, we propose an Extended Manifold Ranking (EMR) algorithm, which subtly combines suppressed/active nodes and mid-level information. Extensive experimental results demonstrate that the proposed algorithm performs favorably against the state-of-art saliency detection methods in terms of different evaluation metrics on several benchmark datasets.

## 1. Introduction

Human visual system can locate the most important regions in a scene effortlessly and accurately. The selective attention mechanism can facilitate the high-level cognitive task. Therefore, saliency detection to model biological visual systems has received increasing interest from computer science, psychology and neurobiology[1] in recent years. As an essential pre-processing step, numerous saliency models have been applied to various computer vision fields, such as image segmentation[2], object recognition[3], visual tracking[4], image/video compression[5] and image retrieval[6], and so on.

Itti et al.[7] believed that human visual system paid more attention to high-contrast regions and computed saliency via local contrast. ased on this theory, contrast prior becomes one of the most principles to be adopted by various kinds of saliency models [8, 9] from either local or global view. For local methods[7], center-surround contrast is used to characterize saliency. Due to lack of global information, these methods can highlight the boundaries of the salient objects while failing to detect internal patches of the salient objects (see Fig.1(c)). On the other hand, global contrast[9], which prefers to exploit the global information for saliency detection, can be able to accurately determine the location of the salient objects. However, the effectiveness of these methods in uniformly detecting the salient objects is limited when background regions have similar appearance resulting from local information being ignored (see Fig.1(d)).

Different from adopting contrast prior, many saliency methods[10, 11, 12] formulate their algorithms based on boundary prior, regarding that it is a high probability for image boundaries to be background. Although these methods perform well in some cases, they also have several drawbacks. Firstly, it is not appropriate to treat all regions on the boundary as background for the reason that sometimes the object may appear on the image boundary (see Fig.1(e)). Secondly, most of these methods are effective in uncomplicated cases, but they still struggle in complex scenes due to the feeble low-level features. For the shortcoming of low-level features, many researchers turned to incorporating high-level feature[13, 14]. Some methods[11, 15] adopt task-driven strategies involving supervised learning on image data with pixel-wise annotations. However, obtaining massive amount of manual label data is very expensive and time consuming. on the other hand, a recent trend is to incorporate high-level features to facilitate detection. A type of high-level representation is the concept of objectness. The problem is that using the objectness value directly to compute saliency may cause unsatisfying results in complex scenes if the objectness score fails to predict the correct object[13, 16].

To this end, we put forward an effective graph-based method for saliency detection. The pipeline of our method is shown in Figure 2. At first, instead of simply using all the four sides of an image as background seeds[10, 11], we introduce selective mechanism based on color scatter to get reliable background seeds and compute the saliency value of each node by finding the shortest path from the corresponding node to the virtual background node, thereby generating the background-based saliency map. To suppress the background noise, the foreground-based saliency map is constructed. Specifically, highly confident compact foreground seeds are generated from background-based saliency map. Similar to the background-based saliency detection, the saliency value of each node is measuring by finding the shortest path from the corresponding node to the virtual foreground node and obtain the foreground-based saliency map. After that, the two saliency maps are integrated by the proposed unified function. Finally, based on manifold ranking, an improved saliency propagation mechanism, which introduces suppressed /active nodes and mid-level mid-level information, is proposed to refine the integrated result.

The main contributions of this paper are as follows:
- We propose a saliency detection algorithm based on a hybrid of divergence background and compact foreground on a novel graph structure, which introduces the background/foreground seed and virtual background/foreground node.
- Reasonably edge weights are defined by considering intrinsic color, spatial and edge information of image, which can effectively characterize the relationship between superpixels and distinguish between foreground and background.
- By taking spatial compactness and rarity of salient objects into consideration, a compact and coherent foreground region can be generated.
- We propose a robust saliency propagation mechanism based on manifold ranking to

refine saliency map. The experimental results demonstrate that it is of high versatility that can improve other methods based propagation when applying our propagation mechanism.

The remainder of this paper is organized as follows. Section II describes the related work on salient object detection. Section III shows the detailed description of our proposed salient object detection. The experimental results and performance evaluation are presents in Section IV. Finally, Section V concludes the paper.

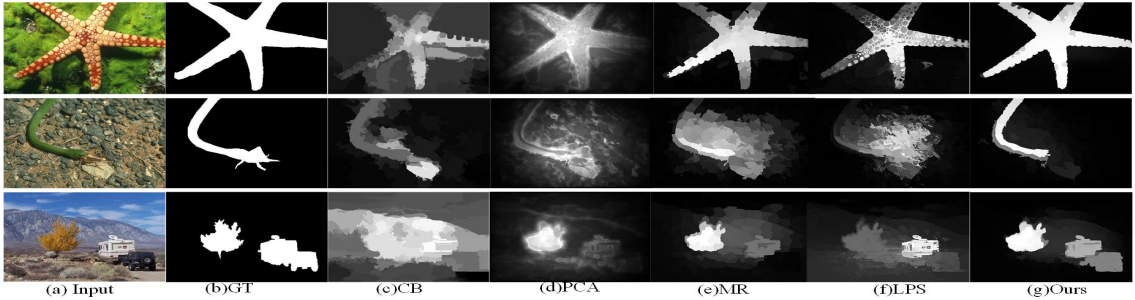

Figure. 1 The differences among various saliency models: including the local contrast based model CB [21], the global contrast based model PCA[57], the background prior based model MR [10], and the objectness based model LPS [33].

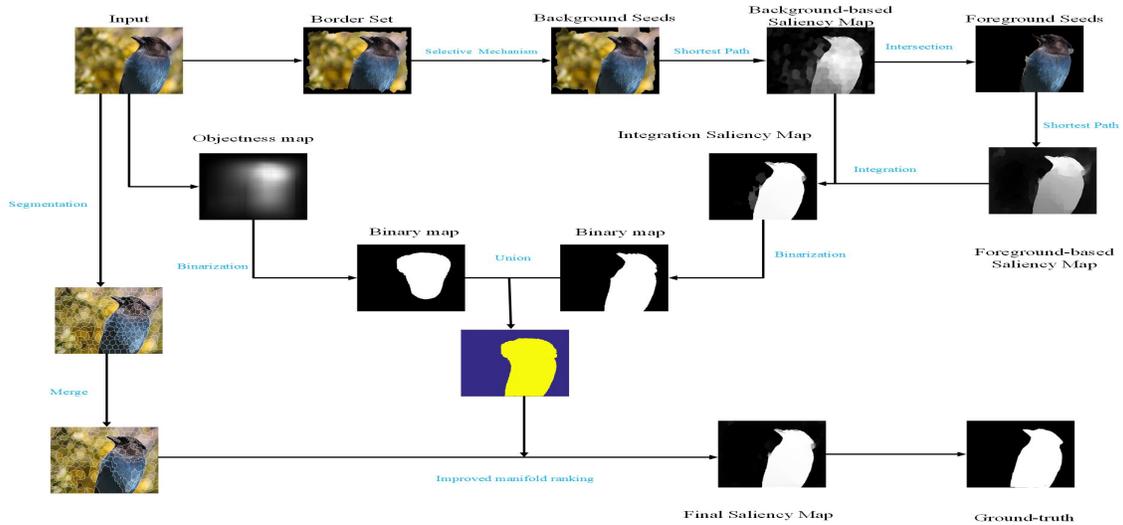

Figure 2 Pipeline of our method.

## 2. Related work

Recently, many prior hypotheses were commonly used in saliency detection. Some early works mainly use the *center prior* assuming the center region of an image with higher probability to be salient object. A Gaussian fall function is usually used to highlight the center regions, and it is either directly combined with other information[8,18], or used as a feature in learning-based methods[11, 19]. *Background prior* is anther widely used principle to compute saliency. Yang et.al[10] computed the saliency of image regions according to the relevance to boundary nodes via manifold ranking. Wei et al[20] computed saliency value of each region by measuring the shortest-path distance to the boundary. In addition, *shape prior*[21], *focussness prior*[22], *segmentation prior*[23] and *sparsity prior*[24] have been also exploited to facilitated the detection of salient objects. In this paper, we propose an method to optimize the image border set based on color scatter and define the robust border regions as background nodes.

A graph can be applied to represent relationships between image elements with affinity measure. Harel et al. [25] proposed the graph based visual saliency (GBVS), which adopted multiple features to extract saliency information on a graph. A hierarchical graph model were developed for saliency detection combining context information[26]. Besides, graph-based methods are often associated with diffusion processes. Thus far, more and more saliency methods have also been proposed by diffusing process to propagation saliency information throughout a graph with different features and affinity measures. Zhang et al.[27] ranked the similarity of image elements with foreground or background cue via graph-based manifold ranking. Li et al.[28] proposed a robust background measure to characterize the spatial layout of an image region with respect to the boundary regions and estimate the saliency via regularized random walks ranking. Jiang et al.[29] formulated saliency detection based on Markov absorption probabilities on an image graph model. Different from preview graph model, we propose a novel graph by introducing background/foreground seed nodes, virtual background/foreground see nodes. Furthermore, a robust saliency propagation mechanism based on manifold ranking is proposed to refine the final saliency map.

In addition, numerous generic object detection methods, which aims at generating the location of objects in an image, have been applied to salient region detection. In [30], a saliency measure was implemented by combining the objectness values of many overlapping windows. Li et al[31] predicted the saliency score of an object candidate by training a random forest model. To improve saliency estimations, Chang et al[13] presented an iterative optimization of energy functions, which combines saliency, objectness and interaction terms. However, the background regions could be further falsely highlighted if the objectness can not measuring the correct object. Recently, a co-transduction algorithm, namely label propagation saliency, is devised for saliency detection via incorporating low-level features and the objectness measure. All of them demonstrate that objectness is helpful to saliency detection. Instead of directly using objectness to compute saliency, we integrate foreground regions based on saleicncy map with objectness to extract reliable supressed nodes and active nodes resulting in boosting salient region detection performance. performance.

## 3. Graph Construction
### *3.1 Pre-processing*

Firstly, we segment the image into $N$ superpixels by the SLIC method and use them as the minimum processing units. Here, we set the number of superpixel nodes $N$=250 in all the experiments. The advantages brought by segmenting an image into several superpixels are that the superpixels can not only capture the structural information of an image but also speed up the processing. Then, an undirected weighted graph G=(*V*, *E*) is constructed. Each nodes $v_i \in V$ corresponds to a superpixel and each edge $e_{ij} \in E$ connects two superpixels. Inspired by [32], we also introduce the concept of virtual nodes. In this paper, we define two virtual nodes, namely, virtual background node $VT_{bp}$ and virtual foreground node $VT_{fp}$. Specifically, we denote that the edge weights between virtual background(foreground) node and well-defined background(foreground) superpixels are assigned with zero.

To improve the detection performance, two extensions are make upon most existing methods[33,34]. First, they connect each node to its spatial neighbors which share a boundary, as well as its spatial neighbors' neighbors which share a boundary with any node in the first set of spatial neighbors. However, each node is only connected to its neighboring nodes in this paper. Furthermore, in order to reduce the geodesic distance of similar superpixels, they define that boundary superpixels are connected to each other. In this graph, we connect the seed nodes to virtual node. The rational behind is that not all boundary superpixels are background.

### *3.2 Construct edge weights of graph*

The edge weights measure the similarity or dissimilarity between nodes. In existing graph-based salient object detection methods, edges are usually weighted by the color distance between the nodes. The drawback of the defined edge weights is that it does not perform well if some part of the distant background has a very similar color to the foreground. Accordingly, we adopt a joint metric taking color difference, spatial distance and edge information into consideration to compute edge weights. It is mainly considered from the following aspects:

(1) According to the cognitive property of color similarity, image regions with similar colors often belong to the same category.

(2) Based on the spatial proximity property, adjacent regions spatially are likely to be of the same label.

(3) In some case, using edge map can be better to highlight the outline between foreground and background than (2) and (3).

Based on above consideration, we can formulate our edge weights:

$$A(i,j) = \exp(-\frac{d_c(i,j) + d_s(i,j) + d_{edge}(i,j)}{2\sigma_w^2}) \quad (1)$$

where $\sigma_w$ controls the strength of weight between a pair of nodes and we set $\sigma_w = 0.1$ empirically. $d_c(i,j)$ is the color difference, $d_s(i,j)$ represents the spatial distance between the nodes *i* and *j*, $d_{edge}(i,j)$ is the intervening contour magnitude[35]. The color difference is defined as :

$$d_c = |c_i - c_j| \quad (2)$$

Here $c_i$ and $c_i$ are the mean value of superpxiel in CIELAB color space, which has been proved to be effective[10]. Existing most methods to compute the spatial distance generally adopt the Euclidean distance, which is prone to cause a large difference especially between opposite image borders owing to their largest distance. However, it is highly possible for image borders to belong to background. To improve the detection performance, we measure the spatial distance by sine spatial distance [36]:

$$d_s(i,j) = \sqrt{(\sin(\pi \cdot |x_i - x_j|))^2 + (\sin(\pi \cdot |y_i - y_j|))^2} \quad (3)$$

where $x_i$ and $y_i$ represent the normalized coordinates of node $i$ respectively. The formulation makes sure that the spatial distance between boundary superpixels especially the nodes at the opposite borders of the image would be small. Refer to [36] for more detail.

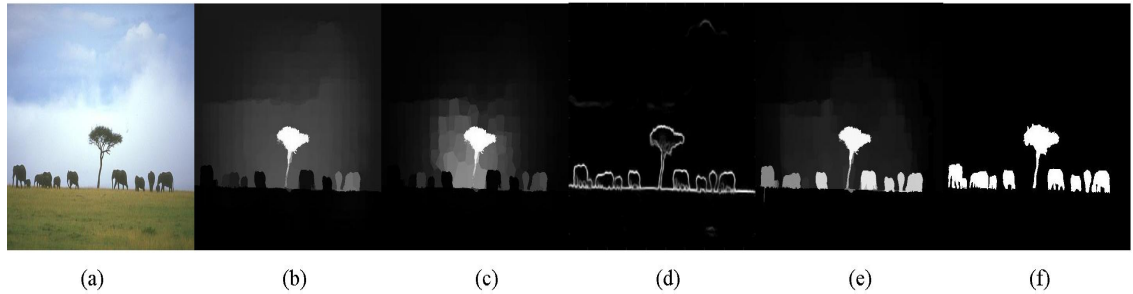

(a)      (b)      (c)      (d)      (e)      (f)

Figure 3. Effects of edge weights. (a) Input; (b) The saliency map by just considering color difference; (c) The saliency map by considering color and spatial information; (c) Edge detection by[55]; (d) The saliency map using our edge weights; (f) Ground-truth.

As for complex scene, it is difficult to distinguish the foreground from the background just based on color feature or both color difference and spatial contrast (as shown in Fig.3(b)-(c)). Fortunately, we observe that it can produce a good result which can pop out the entire contour of the tree and bulls due to texture distinction. Hence, we introduce the intervening image edge cues. We define the intervening contour magnitude

We define the intervening contour magnitude $d_{edge}(i,j)$ as follow:

$$d_{edge}(i,j) = \max_{p \in line(loc_i, loc_j)} E(p) \quad (4)$$

where $loc_i$ represents the centroid location of superpixel $i$, $line(loc_i, loc_j)$ is the line connecting node $i$ and node $j$, $p$ runs over every pixel on the line. $E(p)$ is the corresponding edge probality on an edge map $E$. However, there also exits a problem that $d_{edge}(i,j)$ can be extremely large for the opposite image borders as their strong edges on the line connecting them. In order to alleviate such problem, we set $d_{edge}(i,j)=\beta d_c(i,j)+(1-\beta)ds(i,j)$ for $d_{edge}(i,j)$ among all borders nodes. Here, we set $\beta=0.5$.

*3.3 Salient object probability with background information*

*3.3.1 background seeds acquisition*

The nodes along boundary are usually employed as the background seeds because object is likely to appear at or near the center of an image. Nevertheless, in some cases, the salient object may also appear at the border regions, misleading the saliency value to be 0 if the object is mistaken for background regions (as shown in Fig.4(c)). Consequently, we propose an mechanism based on divergence information to select robust background regions from the border regions.

The background regions present such situation that they usually have diverse appearances and widely distributed. Based on the above findings, the color and texture scatter degree are proposed to compute the compactness map [58]. Different from [58], we extract background seeds via divergence information, which is based on edge weights and center prior. According to edge weights, we define the divergence of superpixel $i$ as

For a superpixel $i$, color scatter can be computed by the following reciprocal form of weight:

$$Div_c(i) = normalize(\sum_{j=1}^{N} a_{ij} \cdot |s_j - \rho_i|) \quad (5)$$

and

$$\rho_i = normalize(\sum_{j=1}^{N} a_{ij} s_j) \quad (6)$$

Here, $s_j$ is the position of superpixel $j$; $\rho_i$ indicates the mean position of superpixel $i$, which has been subjected to weighting; and $N$ is the number of superpixels; *normalize(x)* is a function that normalizes $x$. Considering that image center region is more likely to be salient object, we introduce the center prior guided divergence information, which is defined in the following formulation

$$Div_m(i) = normalize(\sum_{j=1}^{N} a_{ij} \cdot |s_j - M|) \quad (7)$$

where $M$ is the normalized spatial coordinate of the image center. Then, the divergence value of superpixel $i$ is finally measured using

$$Div(i) = normalize\left(Div_c(i) + Div_m(i)\right). \quad (8)$$

In this formulation, the larger the *Div(i)* is, the more the corresponding superpixel $i$ is likely to be the background.

TABLE I: Probability distribution on different datasets.

| Dataset | Top(%) | Down(%) | Left(%) | Right(%) |
|---|---|---|---|---|
| ASD | 0.2 | 1.6 | 0.3 | 0.5 |
| ASD | 5.3 | 22.4 | 6.8 | 7 |
| DUT | 2.48 | 14.4 | 5.2 | 4.3 |

Based on the divergence information, we can remove the unreliable background regions in border set by threshold value. As shown in TABLE I, we observer that there has different probabilities of the object appearing at different boundaries. For example, the probability of the object connecting with the down side is larger than those of three sides. If we adopt the same threshold value for all border sides, the results may be not very accurate. Accordingly, we choose different threshold values for different sides instead of setting a single value. In specific, the superpixels are removed from the border sets if their divergence values are lower than the mean divergence over the entire map for down side. For top side, we set the threshold value is one third of the down side's, which is also twice for the left and right sides's for simplicity. Finally, we can get the robust background seeds. As shown in Fig.4(d), our method can suppress effectively such a situation that the object on the border is mistaken as background.

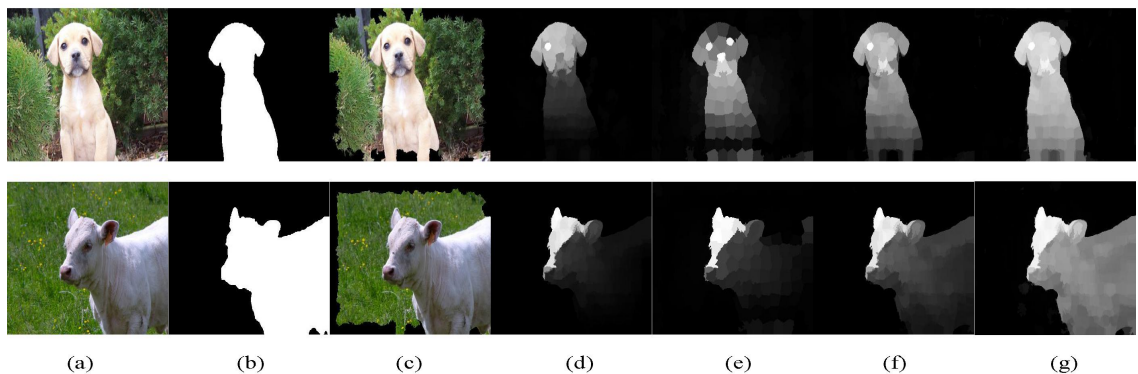

(a)     (b)     (c)     (d)     (e)     (f)     (g)

Figure 4 From left to right: (a) Input; (b) Ground-truth; (c) Our background regions; (d) Use all boundary nodes based on our Eq.6; (e) Using all boundary nodes based on manifold ranking; (f) Use our background nodes based on manifold ranking; (g) Our background-based saliency map.

### 3.3.2 background-based saliency map

Based on the graph G with the above defined edge weights and the well-defined background seeds, the confidence of each superpixel $i$ as background is measured by accumulating the edges weights of the shortest path from $i$ to the virtual background node $VT_{bp}$:

$$con_{bp}(i) = \min_{u_1=i,\ldots,u_k=VT_{bp}} \sum_{j=1}^{k} A(u_m, u_{m+1}) \quad (9)$$

The magnitude of $con_{bp}(i)$ represents the similarity between superpixel $i$ and the well-defined background region, in other words, the higher $con_{bp}(i)$ is, the higher the probability of superpixel $i$ belonging to foreground is. For a node representing a background superpixel, the confidence of superpixel $i$ is normally lower resulting from the shorter of the path to the virtual background node and the smaller of the edge weights over the shortest path. Similarly, it is usually larger for a node representing a foreground node. As shown in Fig.4(e), the method of the shortest path to the virtual node with the help of the select background nodes works well.

## 3.4 Salient object probability with foreground information

### 3.4.1 foreground seeds

While the background-based saliency map can pop out the object well whereas in some cases due to some true background superpixels not being involved in the background seeds and the faint difference between background and foreground (as shown in Fig.4(e)). Therefore, we exploit foreground seeds to construct foreground-based map to inhibit background noises.

We acquire foreground regions based on background-based saliency map. Different from [37] which simply binarizes the saliency map using as adaptive threshold and selects the superprixels whose background-based saliency values are larger than the threshold as foreground seeds. [17] takes the spatial compactness of salient objects into consideration and uses the parametric maxflow to yield the foreground regions. Visual rarity captures the fact that human eyes are often attracted to the rare features in an image but not to the common features. Based on [17], we introduce a regularization term *Rare*. The foreground region can be obtained by solving[38]:

$$fg^* = \arg\min_{fg} \sum_{i=1}^{N}\{(-\ln S_i + \eta area_i) \cdot fg_i + \text{Rare}(i)\} + \sum_{1 \leq i < j \leq N} A_{ij} fg_i fg_j \quad (10)$$

where $area_i$ denotes the area of superpixel $i$. $fg_i \in \{0,1\}$ indicates whether the corresponding superpixel $i$ belong to foreground region. $Rare(i)$ indicates visual rarity of superpixel $i$. According to the visual rarity, image background tends to occupy more area than salient foreground in an image and salient foreground tends to have rare color features. Based on above considerations, visual rarity of superpixel $i$ can be defined as:

$$Rare(i) = normal(\sum_{j=1}^{N} A^1(i,j) + \sum_{j=1}^{N} A^2(i,j)) \quad (11)$$

$$st. A^1(i,j) = \begin{cases} A(i,j) & j \in N_i \\ 0 & otherwise \end{cases}$$

$$A^2(i,j) = \begin{cases} A(i,j) & d_c(i,j) < \phi \\ 0 & otherwise \end{cases}$$

where $N_i$ denotes the spatial neighbors of node $i$, $d_c(i,j)$ denotes the color distance between the nodes $i$ and $j$. $\phi$ is a constant and we set it as 0.15 in our method. As shown in Figure 5(e-f), our method can product more compact foreground regions than those by adaptive threshold or using Eq.10 without introducing rare term. In addition, our results can more accurately pop out the salient objects.

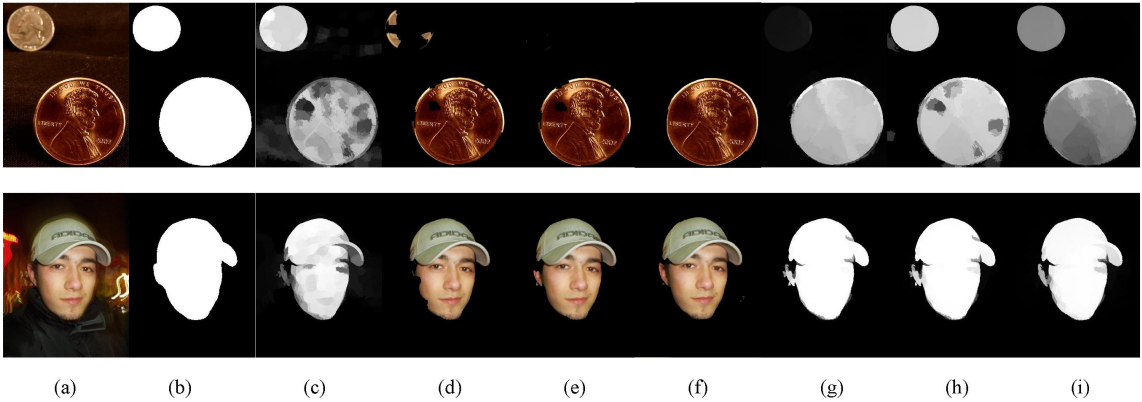

(a) (b) (c) (d) (e) (f) (g) (h) (i)

Figure 5 From left to right: (a) Input; (b) Ground-truth; (c) Background-based saliency map; (d) Foreground regions generated by adaptive threshold; (e) Foreground regions generated without introducing rare; (f) Foreground regions generated by the proposed method; (g) Foreground-based saliency map; (h) Integration saliency map; (i) Refinement saliency map.

### 3.4.2 foreground-based saliency map

Similarly, we can also compute the confidence of each superpixel $i$ as foreground, which is measured by accumulating the edges weights of the shortest path from $i$ to the virtual foreground node $VT_{fp}$:

$$con_{fp}(i) = 1 - \min_{u_1=i,\ldots,u_k=VT_{fp}} \sum_{j=1}^{k} A(u_m, u_{m+1}) \quad (12)$$

If the $con_{fp}(i)$ is lower, it represents a lower difference between $i$ and the reliable foreground regions, which also means that it has higher confidence to be foreground. As shown in Fig.5(g), the undesired highlighted background regions in the background-based saliency map are greatly suppressed in the foreground-based one owing to the contrast to the foreground seeds.

### 3.5 Integration and refinement

### 3.5.1 Integration

We can find that the salient object can be uniformly highlighted in the background-based saliency map including some noises while the foreground-based saliency map can better inhibit the noises exactly. This suggests that a complementary combination of them could generate better results. Based on [37], we use the following formulation to incorporate them:

$$S_{com}(i) = con_{bp}(i) * (1 - \exp(-\kappa * con_{fp}(i))) \quad (13)$$

Where κ is the balancing factor between these two saliency maps, which is set to 4 in our experiments. As shown in Fig.5(h), the integrated saliency map can not only highlight the salient object uniformly but also significantly restrain the noises.

### 3.5.2 saliency propagation based on improved manifold ranking algorithm

The coarse saliency maps need to be further optimized by saliency propagation whose task is to reliably and accurately transmit saliency values from the labeled superpixels to the remaining unlabeled ones. It is noted that we directly use the rough saliency value of each superpixel to replace the binary queries in the original manifold ranking which will introduce errors in the binarization.

Accordingly, two extensions are made upon manifold ranking algorithm. First, we introduce the suppressed nodes and active nodes into the data manifold. The suppressed nodes will never disseminate information to the neighboring nodes during diffusion, and active nodes otherwise. $I_f$=diag$\{\delta_1,...,\delta_N\}$, where $\delta_i$=0 if the corresponding superpixel $i$ is a suppressed node and an active node is represented by $\delta_i$=1. Different from [39], which defined so-called sink point mainly based on center weight, we adopt objectness map[56] and integration map to determine suppress nodes and active nodes. First, the higher threshold of the two-level Ostu's adaptive threshold method[40] is performed on Objectness map(OM) and the integration map(Com), to respectively obtain $fg_{om}$ and $fg_{Com}$. Then, the union set $fg_T=fg_{om} \cup fg_{Com}$ is highly confident object regions, in other words, the regions in the complement of $fg_T$ are more likely to be background. Therefore, we set $\delta_i$=1 if the superpixel $i \in fg_T$ and 0 otherwise.

In addition, the appearance of manifold ranking is based on the hypothesis that the background has a high contrast with objects, which determines that it is difficult to imprecisely pop out salient objects when the background is complex and the difference between foreground and background in an image is low. Accordingly, we merge a quantity of superpixels together into much bigger regions using the mid-level clustering algorithm [41]. As a result, the nodes with the same clusters would have the similar value. Therefore, the mid-level similarity matrix $P$ can be defined as:

$$p_{ij} = w_{ij} + q_{ij} \quad (14)$$

With

$$q_{ij} = \begin{cases} 1 & i \text{ and } j \text{ are in the same subregion} \\ 0 & otherwise \end{cases} \quad (15)$$

$$w_{ij} = \begin{cases} a_{ij} & j \in N_i \\ 0 & otherwise \end{cases} \quad (16)$$

Based on above definitions, the optimal ranking of queries are computed by solving the following optimization problem:

$$f^* = \arg\min_f \frac{1}{2}(\sum_{i,j=1}^{N} p_{ij} \| \frac{\delta_i f_i}{\sqrt{d_{ii}}} - \frac{\delta_j f_j}{\sqrt{d_{jj}}} \|^2 + \mu \sum_{i=1}^{N} \| \delta_i f_i - y \|^2) \quad (17)$$

where the parameter $\mu$(setting as 0.01 by empirically) controls the balance of the smoothness constraint (the first term) and the fitting constraint (the second term). The minimum solution is computed by setting the derivative of the above function to be zero. The resulted ranking function can be written as:

$$f^* = (I - \alpha T I_f)^{-1} y \quad (18)$$

where I is an identity matrix, α = 1/(1 + μ) and S is the normalized Laplacian matrix, T = $D^{-1/2}PD^{-1/2}$. We can get another ranking function by using the unnormalized Laplacian matrix in Eq. 18 inspired by [10] which can achieve better performance:

$$f^* = (D - \alpha P I_f)^{-1} y \quad (19)$$

the saliency of each node is defined as its ranking score computed by Eq. 15 which is rewritten as

$$f^* = BY \quad (20)$$

The learnt optimal affinity matrix B is equal to $(D-\alpha PI_f)^{-1}$. The effectiveness of our refinement is illustrate in Figure 5(i). Our method is able to detect the foreground uniformly and extract the well-defined object boundary.

4. Experiment results
4.1 Dataset and evaluation metrics
4.1.1 Dataset

To evaluate our saliency detection algorithm, we conduct a series of experiments on four benchmark datasets with pixel-wise manually labeled ground truths.

(1) ASD[42]: It includes 1000 images selecting from the MSRA salienct object database[19], which covers a large variety of scenarios. This dataset have usually only one salient object and there are strong contrast between backgrounds and foregrounds.

(2) DUT-OMRON[10]: It consist of 5168 images natural images carefully labeled by five users. So far, none of existing saliency methods can achieve a high accuracy on this dataset.

(3) ECSSD[18]: It contains 1000 semantically meaningful and structurally complex images acquired from the Internet with pixel-level saliency labeling.

(4) SOD[43]: It consists of 300 images selected from the Berkeley Segmenttation dataset with labeled ground truth. The dataset are pretty difficult with different biases such as number of salient objects, image clutter and center-biases and so on.

4.1.2 Evaluation metrics

To compare the performance, we first use standard precision-recall(PR) curves to evaluate the performance of our method. For each continuous saliency map can be converted into a binary map by segmenting it with a threshold varying from 0 to 255[42]. And then, A pair of precision and recall values can be generated when the binary map is compared with the ground truth. A PR curve is then obtained by varying the threshold from 0 to 1.

Fore comprehensively assessing the salient object detection model, we also compute

the maximal F-measure, which is a harmonic mean of precision and recall. Here the F-measure is defined as:

$$F_\beta = \frac{(1+\beta^2) \cdot \text{Precision} \times \text{Recall}}{\beta^2 \times \text{Precision} + \text{Recall}} \quad (21)$$

where $\beta^2$ is set to 0.3 to emphasize precision[9].

Although commonly used, PR curves have limited value because they fail to consider true negative pixels. For a more balanced comparisons, we introduce the mean absolute error(MAE), which is defined as the average pixelwise absolute difference between the ground truth(G) and saliency map(S)[44], as another evaluation criterion:

$$MAE = \frac{1}{W \times H} \sum_{x=1}^{W} \sum_{y=1}^{H} |S(x,y) - G(x,y)| \quad (22)$$

where $W$ and $H$ are the width and height of the saliency map $S$. MAE is mainly to measure the numerical distance between the ground truth and the estimated saliency map, and is more meaningful in evaluating the applicability of a saliency model in a task such as object segmentation.

**4.2 Comparison with state-of-the-art**

In this paper, we extensively present comparison of the proposed algorithm against thirteen state-of-the-art saliency detection methods including GS [20], MR [10], BFS [37], BL [48], BSCA [49], LPS [33], MB [47], NCU [34], RR [28], SP [12], SRD [54], SBD [53], MILPS [59] on the ASD, DUT-OMRON, SOD, and ECSSD datasets. For fair evaluation, we either directly use the results provided by the original authors or run their own implementations through the source codes publicly available online.

**4.2.1 Evaluation on ASD dataset**

Fig.6 shows the PR curves, F-measure, AUC and MAE on ASD dataset. As shown in Fig.6(a) and Fig.6(b), the PR curves clearly demonstrate that our model outperforms the other algorithms. our model is not only better than the GS [20], MB [47], NCU [34] methods which are top-performing methods for saliency detection in recent study, but also outperforms the BFS [37], BSCA [49], which are also based on background priors. The average precision, recall, F-measure, AUC and MAE are presented the first row in the bar graph Fig.6(c)-Fig.6(d). In the bar graph, our algorithm performs the second best in terms of MAE, which is slightly worse than SBD [53], but the highest precision, F-measure and AUC of 0.9513, 0.934, 0.88699 are accomplished by our model outperforming other 13methods. Note that the MR [10] has large precision value, since they tend to detect the most salient regions at the expense of low recall while LPS [33] and RR [28] also show the same imbalance.

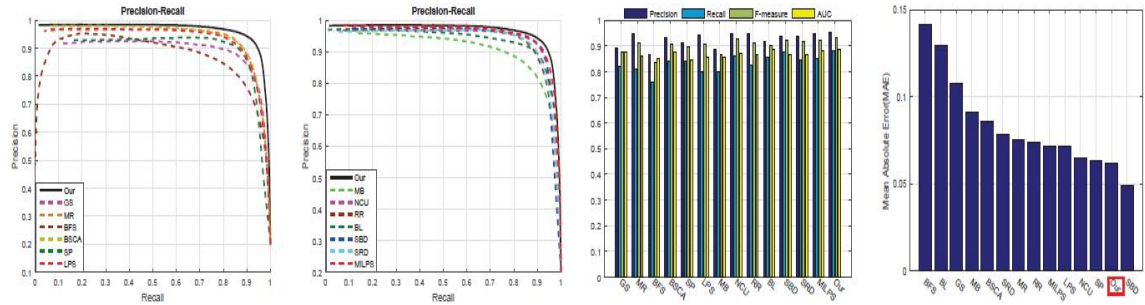

Figure 6. The comparison results of the proposed algorithm with the state-of-the-art methods on the ASD dataset.

### 4.2.2 Evaluation on DUT-OMORN dataset

Though the images in ASD dataset contain different kinds of content, the background of an image is usually relatively simple. Hence, we bring in the DUT-OMORN dataset, which has relatively complex background and is more challenging than ASD dataset. Fig.7 reports the performance comparison on DUT-OMORN dataset. As this is a challenging dataset, the performance significantly degrades in terms of all metrics. The PR curve of the proposed method is superior to other methods except for SBD [53], which is just higher at low recall rates and the curve drops sharply at higher recall rates. As reported in Fig.7(c)-(d), our model obtains the second and third best in F-measure and MAE, while the AUC is the best. Although the MR [10] has large precision value, it still suffers from low recall and high MAE. These results validate the strong robustness and capacity of our method.

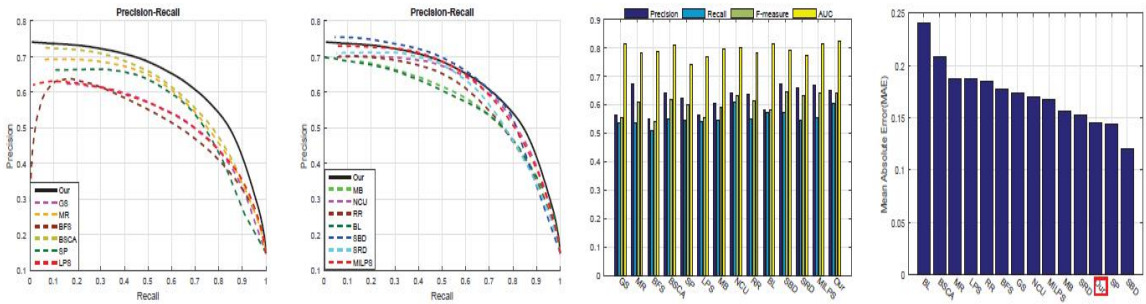

Figure 7 The comparison results of the proposed algorithm with the state-of-the-art methods on the DUT-OMORN dataset.

### 4.2.5 Evaluation on SOD and ECSSD datasets

Although DUT-OMORN dataset has relatively complex background, it usually has only one salient object in an image. To reliably compare how well various approaches detect salient regions in multiple-object and complex scenes, experiments are performed on SOD and ECSSD datasets. Fig.8 and Fig.9 display the evaluation results of the proposed method compared with other 13 approaches.

On SOD (Fig.8), our model achieves the best performance in terms of F-measure and MAE, while BL \cite{50} obtains the best AUC curve (Fig.8(c)-(d)). On the other hand, the PR curve of our method is the best one among those methods. As observed from Fig.8(a)-(b), our method gets the highest precision value in the whole recall interval

[0,1], where the highest value can reach up to 0.7142.

On ECSSD (Fig.9), our model performs the second best in terms of MAE, with a very minor margin(0.003) to the best, but our algorithm have the highest precision, F-measure and AUC of 0.8134, 0.7765, 0.82008. In PR curves (Fig.9), our method is superior to those competitive methods.

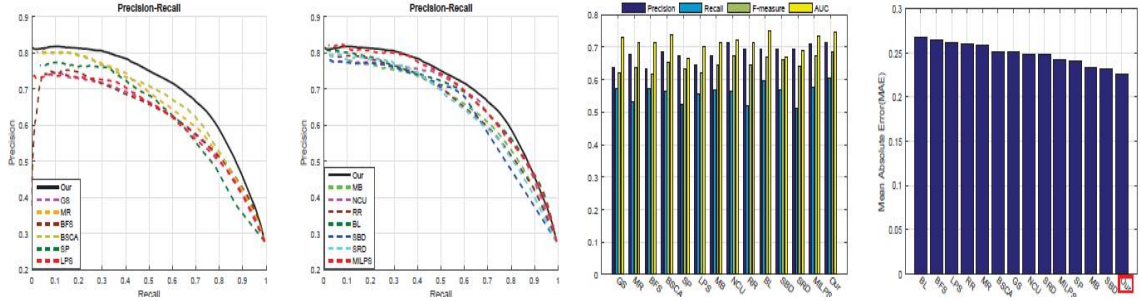

Figure 8. The comparison results of the proposed algorithm with the state-of-the-art methods on the SOD dataset.

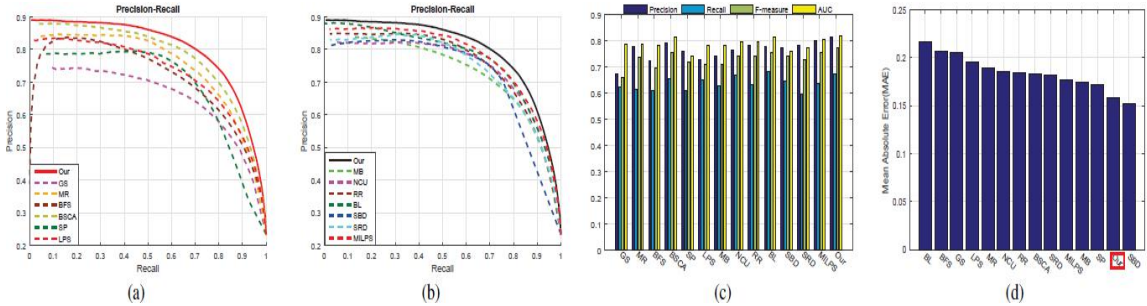

Figure 9. The comparison results of the proposed algorithm with the state-of-the-art methods on the ECSSD dataset.

**4.2.5 Visual comparison**

Figure 10-13 show some visual comparisons of the best methods on the four datasets. For simple background and single-object images shown in Figure 10, our method can uniformly highlights the salient object with few scattered patches. In addition, we can observe that the proposed method can also effectively handle the challenging cases where the background is complex or very similar to the foreground. For example, as shown in Figure 11 and Figure 13, the other methods often fail to distinguish the background and foreground regions with very similar color while our approach performs well. Due to the reasonable edge weights combining with robustness background and foreground seeds, our algorithm can effectively assign different value to the dissimilar nodes resulting in separating them successfully. For images with multiple objects shown in Figure 12, most saliency detection methods often tend to miss patches of objects or incorrectly include background noise into the final results. By contrast, our method have a good ability to pop out all the salient objects successfully. The introduce of improved manifold ranking algorithm makes can further improve the effect, thus detecting more accurate salient objects even from low contrast foreground and cluttered background.

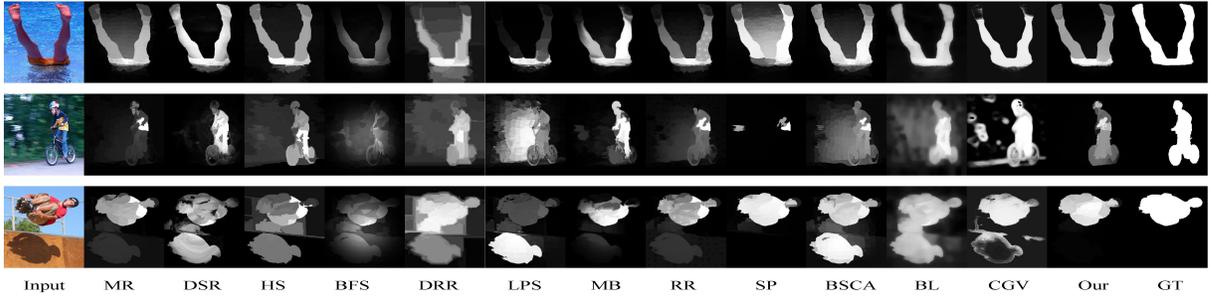

Figure 10. Visual comparison among different methods on ASD dataset.

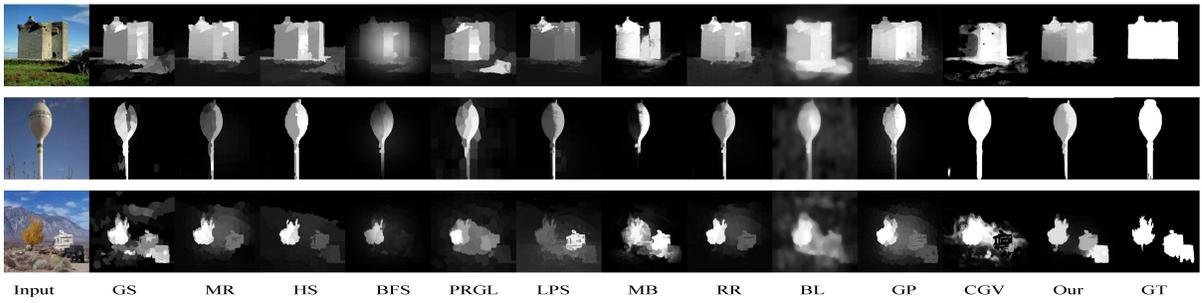

Figure 11. Visual comparison among different methods on DUT dataset.

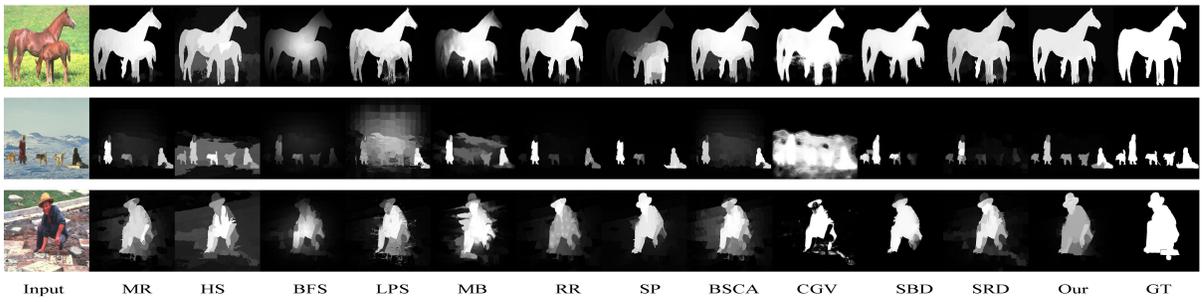

Figure 12. Visual comparison among different methods on SOD dataset.

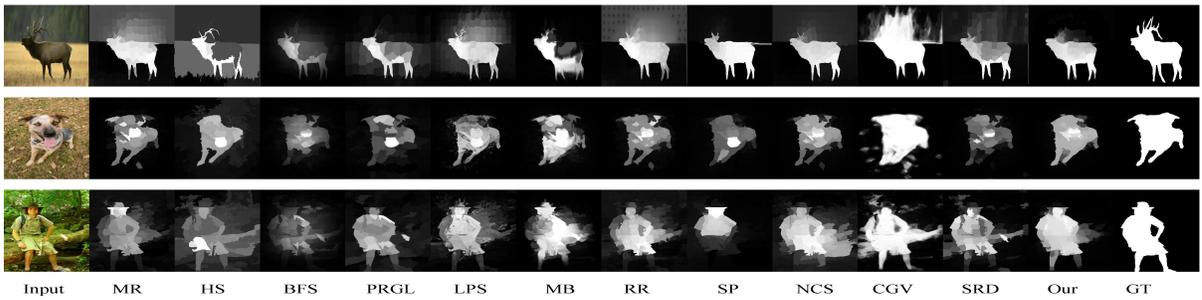

Figure 13. Visual comparison among different methods on ECSSD dataset.

## 4.3 Validation of the proposed approach

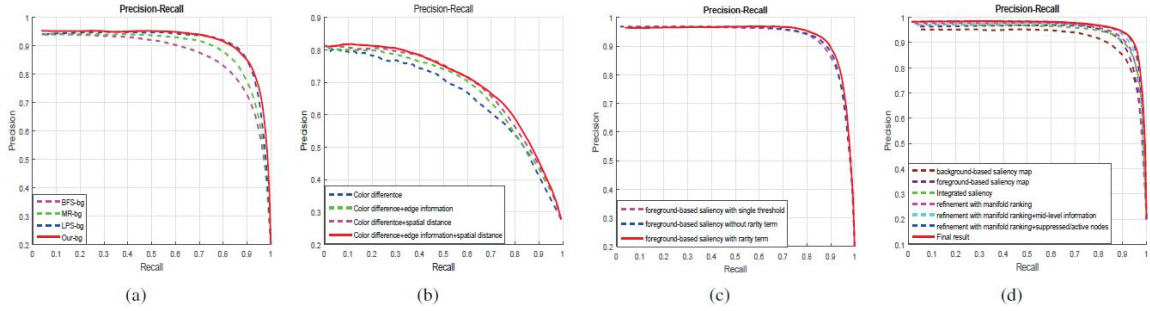

Figure 14. PR curves with different design options of the proposed method.

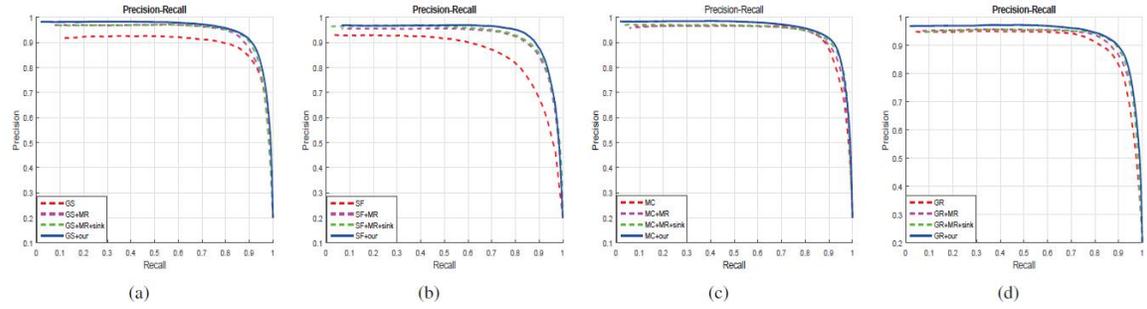

Figure 15. PR curves of different methods and their optimized version by different approaches

To demonstrate the effectiveness of our background-based saliency map, we compute the PR curves for background-based saliency map on ASD dataset. We use three methods, i.e., BFS [37], MR [10], LPS [33], to get different background seeds and construct corresponding background-based saliency maps using our shortest path method with different background seeds. The resulting curves are shown in Fig.14(a). The magenta line provides the performance of the method that constructs saliency map using the background seeds of BFS [37], the green line gives the performance of the method that generates saliency map using the background seeds of MR [10], the blue line shows the performance of the method that generates saliency map using the background seeds of LPS [33] and the red line presents the result using our method. As shown by these curves, our method outperforms those competitive methods, which illustrates the effectiveness of our background-based saliency map.

In order to illustrate the effectiveness of our edge weights, we compute PR curves for the saliency maps with different measures on SOD dataset. Since we use color difference, spatial distance and edge information for edge weights, hence we process the edge weights with different computation methods. The resulting curves are presented in Fig.14(b). The green curve in Fig.14(b) supplies the performance of the method using both color difference and edge information while the magenta curve provides the performance of the method using both color difference and spatial distance. These two curves demonstrate that both spatial distance and edge information also make contributions besides the color distance. The black curve provides the performance of the method using our edge weights, which demonstrates that our method is able to better detect salient objects in images than other methods using edge weights with different forms.

To evaluate the effectiveness of our foreground-based saliency map, we also compute the PR curves for foreground-based saliency map on ASD dataset. As reported in Fig.14(c), the magenta curve provides the performance of the method with a single threshold. Similarly, the blue curve gives the performance of the method without the rarity term, and the red curve presents the performance of the menthod with rarity term. The above curves demonstrate that our method is helpful to detect salient objects.

Since we refine the saliency map based on the EMR algorithm, we evaluate the contributions of separate components in EMR algorithm. Fig.14(d) shows the PR curves for various saliency methods on ASD dataset. The green PR curve represents the result of the integrated saliency. The magenta PR curve gives the performance of the method refined with manifold ranking. The cyan curve provides the effect of the method refined with mid-level information. The blue curve shows the effect of the method refined with suppressed / actived nodes. The red curve presents our final result. Furthermore, in order to demonstrate that our refinement method is of high versatility that can improve other methods based propagation when applying our propagation mechanism, we apply our refinement mechanism to optimize state-of-the-art results on ASD dataset, including GS [20], SF [44], MC [60], GR [61]. From Fig.15(a)-(d), we can see that all of them are significantly improved to a similar level after our method. In addition, we can also find that our performance is better than those optimized by manifold ranking [10] (MR) or the method introducing sink points into manifold ranking (MR+sink) [39]. Based on the above observations, our refinement contributes to the overall performance.

**4.4 Running time**

The Running time test is conducted on a 64-bit PC with Intel Core i5-4460 CPU @ 3.20GHz and 8GB RAM. All the tested codes are provided by the authors and run unchanged in MATLAB R2015a with some C++ mex implementations. Average running time is computed on the ASD dataset. We choose several competitive accuracy methods or those akin to ours, and the results are shown in Table 2. The proposed algorithm is significantly faster than LPS, RR and BFS; and although being slower than MR and GS, our method still outperforms them both considering the overall evaluation performances.

Table II. Running time test results (seconds per image).

| Method  | GS    | LPS   | MR    | RR   | BFS   | Our   |
|---------|-------|-------|-------|------|-------|-------|
| Time(s) | 0.425 | 3.376 | 0.715 | 3.56 | 7.513 | 1.043 |

**4.5 Limitation and analysis**

Although our method can perform well in most cases, it still cannot accurately extract the complete salient objects in some challenging scenarios. As shown in Figure 16, if salient object regions are not effectively highlighted or background regions are falsely regarded as foreground in both background-based saliency map and foreground-based saliency map, it is difficult for our method to achieve the satisfactory detection result.

Since we mainly use the low-level and mid-level feature, we will investigate efficient methods that incorporate high-level feature with the help of deep learning to achieve a better performance in our future work.

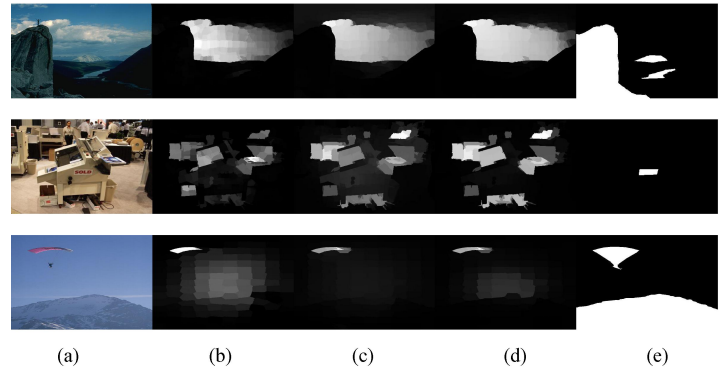

(a)　　　　　(b)　　　　　(c)　　　　　(d)　　　　　(e)

Figure 16. Some failure examples. (a) Input. (b) background-based saliency maps. (c) foreground-based saliency map. (d) Final saliency maps. (e) Ground-truth.

## 5. Conclusions

This paper propose a novel graph structure for salient object detection in which both divergence background and compact foreground are utilized. First, we construct a novel graph and define edge weights, which consider color, spatial and edge information. Then we acquire the robust background nodes based on background color divergence and compute the background-based saliency map. Furthermore, we get the reliable foreground regions based on background-based saliency map and generate the foreground-based saliency map. Finally, we integrate background-based saliency map and foreground-based saliency map to obtain the final saliency map smoothing by our optimization mechanism. Experimental results demonstrate the highly efficiency of our method that performs much better than those state-of-the-art methods.